\documentclass[11pt,a4paper]{article}

\usepackage{times,latexsym}
\usepackage{url}
\usepackage[T1]{fontenc}

\usepackage[acceptedWithA]{tacl2018v2}

\usepackage{color}
\usepackage{hhline}
\usepackage{amsmath}
\usepackage{multirow}
\usepackage{booktabs}
\usepackage{colortbl}
\usepackage{graphicx}
\usepackage{tabularx}
\usepackage{arydshln}
\usepackage{enumitem}

\definecolor{gray}{rgb}{0.89, 0.89, 0.89}

\newcommand{\specialcell}[2][c]{\begin{tabular}[#1]{@{}c@{}}#2\end{tabular}}
\newcommand{\specialcellleft}[2][l]{\begin{tabular}[#1]{@{}l@{}}#2\end{tabular}}
\newcommand{\smallsc}[1]{\fontsize{8}{9}{\textsc{#1}}} 
\newcommand{\smalltt}[1]{\fontsize{8}{9}{\texttt{#1}}}

\usepackage{xspace,mfirstuc,tabulary}

\newif\iftaclinstructions
\taclinstructionsfalse 
\iftaclinstructions
\renewcommand{\confidential}{}
\renewcommand{\anonsubtext}{(No author info supplied here, for consistency with
TACL-submission anonymization requirements)}
\newcommand{\instr}
\fi

\iftaclpubformat 

\else

\fi

\title{Still a Pain in the Neck:\\
Evaluating Text Representations on Lexical Composition}

 \author{Vered Shwartz ~~~~~~~~~~~~~~~~~~~~~~~~~~~~~~~~~~~~~~~ Ido Dagan\\
  Computer Science Department, Bar-Ilan University, Ramat-Gan, Israel\\
  {\tt vered1986@gmail.com	~~~~~~	\tt dagan@cs.biu.ac.il }}

\date{}

\begin{document}
\maketitle
\begin{abstract}
Building meaningful phrase representations is challenging because phrase meanings are not simply the sum of their constituent meanings. Lexical composition can shift the meanings of the constituent words and introduce implicit information. We tested a broad range of textual representations for their capacity to address these issues. We found that, as expected, contextualized word representations perform better than static word embeddings, more so on detecting meaning shift than in recovering implicit information, in which their performance is still far from that of humans. Our evaluation suite, consisting of six tasks related to lexical composition effects, can serve future research aiming to improve representations.  
\end{abstract}

\section{Introduction}
\label{sec:intro}
Modeling the meaning of phrases involves addressing semantic phenomena that pose non-trivial challenges for common text representations, which derive a phrase representation from those of its constituent words. One such phenomenon is \emph{meaning shift}, which happens when the meaning of the phrase departs from the meanings of its constituent words. This is especially common among verb-particle constructions (\textit{carry on}), idiomatic noun compounds (\textit{guilt trip}) and other multi-word expressions (MWE, lexical units that form a distinct concept), making them ``a pain in the neck'' for NLP applications \cite{sag2002multiword}. 

A second phenomenon is common for both MWEs and free word combinations such as noun compounds and adjective-noun compositions. It happens when the composition introduces an \emph{implicit meaning} that often requires world knowledge to uncover. For example, that \textit{hot} refers to the \emph{temperature} of \textit{tea} but to the \emph{manner} of \textit{debate} \cite{hartung2015distributional}, or that \textit{olive oil} is made \emph{of} olives while \textit{baby oil} is made \emph{for} babies \cite{shwartz-waterson:2018:N18-2}. 

There has been a line of attempts to learn compositional phrase representations \cite[e.g.][]{mitchell2010composition,baroni-zamparelli:2010:EMNLP,wieting-mallinson-gimpel:2017:EMNLP2017,poliak-EtAl:2017:EACLshort}, but many of these are tailored to a specific type of phrase or to a fixed number of constituent words, and they all disregard the surrounding context. Recently, contextualized word representations boasted dramatic performance improvements on a range of NLP tasks \cite{peters-EtAl:2018:N18-1,radford2018improving,devlin2018bert}. Such models serve as a function for computing word representations in a given context, making them potentially more capable to address meaning shift. These models were shown to capture some world knowledge \cite[e.g][]{zellers-EtAl:2018:EMNLP}, which may potentially help with uncovering implicit information. 

In this paper we test how well various text representations address these composition-related phenomena. Methodologically, we follow recent work that applied ``black-box'' testing to assess various capacities of distributed representations \cite[e.g.][]{adi2016fine,conneau-EtAl:2018:Long}. We construct an evaluation suite with six tasks related to the above two phenomena, as shown in Figure~\ref{fig:map}, and develop generic models that rely on pre-trained representations. We test six representations, including static word embeddings \cite{mikolov2013efficient,pennington2014glove,bojanowski2017enriching} and contextualized word embeddings \cite{peters-EtAl:2018:N18-1,radford2018improving,devlin2018bert}. Our contributions are as follows:

\begin{enumerate}[noitemsep,leftmargin=*]
    \item We created a unified framework that tests the capacity of representations to address lexical composition via classification tasks, focusing on detecting meaning shift and recovering implicit meaning. We test six representations and provide an in depth analysis of the results.
    \item We relied on existing annotated datasets used in various tasks, and re-casted them to fit our classification framework. We additionally annotated a sample from each test set to confirm the data validity and estimate the human performance on each task. 
    \item We provide the classification framework, including data and code and available at \url{https://github.com/vered1986/lexcomp}, which would allow testing future models for their capacity to address lexical composition. 
\end{enumerate}

Our results confirm that the contextualized word embeddings perform better than the static ones. In particular, we show that indeed modeling context contributes to recognizing meaning shift: on such tasks, contextualized models performed on par with humans. 

Conversely, despite hopes of filling missing information with world knowledge provided by the contextualized representations, the signal they yield for recovering implicit information is much weaker, and the gap between the best performing model and the human performance on such tasks remains substantial. We expect that improving the ability of such representations to reveal implicit meaning would require more than a language model training objective. In particular, one future direction is a richer training objective that simultaneously models multiple co-occurrences of the constituent words across different texts, as is commonly done in noun compound interpretation \cite[e.g][]{oseaghdha-copestake:2009:EACL,shwartz-waterson:2018:N18-2,P18-1111}. 

\begin{figure}[!t]
    \scriptsize
    \begin{tabular}{p{0.001em} c|>{\centering\arraybackslash}p{2.255cm} p{0.001em} >{\centering\arraybackslash}p{1.7cm} p{0.001em} c}
\multirow{6}{*}{\rotatebox{90}{\parbox[c][0.01em][c]{3cm}{\bfseries\centering phrase type}}}
& \multirow{2}{*}{\rotatebox{90}{\parbox[c][0.01em][c]{1cm}{\centering MWE}}} & & | & & | &  \\
& & \textbf{VPC Classification} & | & & | & \textbf{Phrase} \\
& & \textbf{LVC Classification} & | & & | & \textbf{Type} \\
& & & | & & | & \\
\cdashline{3-7}
& & & | & & | & \\
& \multirow{2}{*}{\rotatebox{90}{\parbox[c][0.01em][c]{1.2cm}{\centering Free word}}} & & | & & | & \\
& & \textbf{NC Literality} & | & \textbf{NC Relations} & | & \textbf{Phrase}  \\
& & & | & \textbf{AN Attributes} & | & \textbf{Type} \\
& & & | & & | & \\
\hhline{~------}
& & meaning & | & implicit & | & both \\
& & shift & | & meaning & | &  \\
& & \multicolumn{4}{c}{\textbf{phenomenon}} & \\
\end{tabular}

\caption{A map of our tasks according to type of phrase (MWE/free word combination) and the phenomenon they test (meaning shift/implicit meaning).}
\label{fig:map}
\end{figure}


\section{Composition Tasks}
\label{sec:tasks}
We experimented with six tasks that address the meaning shift and implicit meaning phenomena, summarized in Table~\ref{tab:tasks} and detailed below. 

We rely on existing tasks and datasets, but make substantial changes and augmentations in order to create a uniform framework. First, we cast all tasks as classification tasks. Second, we add sentential contexts where the original datasets annotate the phrases out-of-context, by extracting averaged-length sentences (15-20 words) from English Wikipedia (January 2018 dump) in which the target phrase appears. We assume that the annotation does not depend on the context, an assumption that holds in most cases, judging by the human performance scores (Section~\ref{sec:results}). Finally, we split each dataset to roughly 80\% train and 10\% for each of the validation and test sets, under lexical constraints as detailed for each task. 

\subsection{Recognizing Verb-Particle Constructions}  
\label{sec:vpc}

A verb-particle construction (VPC) consists of a head verb and a particle, typically in the form of an intransitive preposition, which changes the verb's meaning (e.g. \textit{carry on} vs. \textit{carry}). 

\paragraph{Task Definition.} Given a sentence s that includes a verb V followed by a preposition P, the goal is to determine whether it is a VPC or not. 

\paragraph{Data.} We use the dataset of \newcite{tu-roth:2012:STARSEM-SEMEVAL} which consists of 1,348 sentences from the British National Corpus (BNC), each containing a verb V and a preposition P annotated to whether it is a VPC or not. The dataset is focused on 23 different phrasal verbs derived from six of the most frequently used verbs (\textit{take, make, have, get, do, give}), and their combination with common prepositions or particles. To reduce label bias, we split the dataset lexically by verb, i.e. the train, test, and validation sets contain distinct verbs in their V and P combinations. 

\begin{table*}[!t]
    \scriptsize
    \centering
    \begin{tabular}{c >{\centering}m{0.25\textwidth} c c c c c}
\toprule
\textbf{Task} & \textbf{Data Source} & \specialcell{\textbf{Train/val/test}\\\textbf{Size}} & \textbf{Input} & \textbf{Output} & \textbf{Context} \\ 
\midrule
\specialcell{\textbf{VPC Classification}} &  \newcite{tu-roth:2012:STARSEM-SEMEVAL} & 919/209/220  & \specialcell{sentence s\\VP $=$ w$_1$ w$_2$} & \specialcell{is VP a VPC?} & \texttt{O} \\
\midrule
\specialcell{\textbf{LVC Classification}} &  \newcite{tu-roth:2011:MWE} & 1521/258/383  & \specialcell{sentence s\\span $=$ w$_1$ ... w$_k$} & \specialcell{is the span\\an LVC?} & \texttt{O} \\
\midrule
\textbf{NC Literality} & \specialcell{\newcite{reddy-mccarthy-manandhar:2011:IJCNLP-2011} \\ \newcite{tratz2011semantically}} & 2529/323/138  & \specialcell{sentence s\\NC $=$ w$_1$ w$_2$\\ target w $\in \{$w$_1$, w$_2\}$} & \specialcell{is w\\literal in NC?} & \texttt{A} \\
\midrule
\textbf{NC Relations} & \specialcell{SemEval 2013 Task  4\\ \cite{S13-2025}} & 1274/162/130  & \specialcell{sentence s\\NC $=$ w$_1$ w$_2$\\paraphrase p} & \specialcell{does p\\explicate NC?} & \texttt{A} \\
\midrule
\textbf{AN Attributes} & HeiPLAS \cite{hartung2015distributional} & 837/108/106  & \specialcell{sentence s\\AN $=$ w$_1$ w$_2$\\ paraphrase p} & \specialcell{does p describe\\the attribute in AN?} & \texttt{A} \\
\midrule
\textbf{Phrase Type} & \specialcell{STREUSLE\\\cite{schneider-smith:2015:NAACL-HLT}} & 3017/372/376 & sentence s & label per token & \texttt{O} \\
\bottomrule
\end{tabular}

    \caption{A summary of the composition tasks included in our evaluation suite. In the context column, \texttt{O} means the context is part of the original dataset, while \texttt{A} is used for datasets in which the context was added in this work.}
    \label{tab:tasks}
\end{table*}

\subsection{Recognizing Light Verb Constructions}  
\label{sec:lvc}

The meaning of a light verb construction (LVC, e.g. \textit{make a decision}) is mainly derived from its noun object (\textit{decision}), while the meaning of its main verb (\textit{make}) is ``light'' \cite{otto1965modern}. As a rule of thumb, an LVC can be replaced by the verb usage of its direct object noun (\textit{decide}) without changing the meaning of the sentence.

\paragraph{Task Definition.} Given a sentence s that includes a potential light verb construction (``\textit{make an easy decision}''), the goal is to determine whether it is an LVC or not. 

\paragraph{Data.} We use the dataset of \newcite{tu-roth:2011:MWE}, which contains 2,162 sentences from BNC in which a potential light verb construction was found (with the same 6 common verbs as in Section~\ref{sec:vpc}), annotated to whether it is an LVC in a given context or not. We split the dataset lexically by the verb. 

\subsection{Noun Compound Literality}
\label{sec:nc_literality}

\paragraph{Task Definition.} Given a noun compound NC $=$ w$_1$ w$_2$ in a sentence s, and a target word w $\in \{$w$_1$, w$_2\}$, the goal is to determine whether the meaning of w in NC is literal. For instance, \textit{market} has a literal meaning in \textit{flea market} but \textit{flea} does not. 

\paragraph{Data.} We use the dataset of \newcite{reddy-mccarthy-manandhar:2011:IJCNLP-2011} which consists of 90 noun compounds along with human judgments about the literality of each constituent word. Scores are given in a scale of 0-5, 0 being non-literal and 5 being literal. For each noun compound and each of its constituents we consider examples with a score $\ge$ 4 as literal, and $\le$ 2 as non-literal, ignoring the middle range. We obtain 72 literal and 72 non-literal examples.

To increase the dataset size we augment it with literal examples from the \newcite{tratz2011semantically} dataset of noun compound classification. Compounds in this dataset are annotated to the semantic relation that holds between w$_1$ and w$_2$. Most relations (except for \texttt{lexicalized}, which we ignore), define the meaning of NC as some trivial combination of w$_1$ and w$_2$, allowing us to regard both words as literal. This method produces additional 3,061 literal examples.

\paragraph{Task Adaptation.} We add sentential contexts from Wikipedia, keeping up to 10 sentences per example. We downsample from the literal examples to balance the dataset, allowing for a ratio of up to 4 literal to non-literal examples. We split the dataset lexically by head, i.e. if w$_1$ w$_2$ is in one set, there are no w'$_1$ w$_2$ NCs in the other sets.\footnote{We chose to split by head rather than by modifier based on the majority baseline that achieved better performance.}

\subsection{Noun Compound Relations}
\label{sec:nc_relations}

\paragraph{Task Definition.} Given a noun compound NC $=$ w$_1$ w$_2$ in a sentential context s, and a paraphrase p, the goal is to determine whether p describes the semantic relation between w$_1$ and w$_2$ or not. For example, ``\textit{part that makes up body}'' is a valid paraphrase for \textit{body part}, but ``\textit{replacement part bought for body}'' is not.

\paragraph{Data.} We use the data from SemEval 2013 Task 4 \cite{S13-2025}. The dataset consists of 356 noun compounds annotated with 12,446 human-proposed free text paraphrases.  

\paragraph{Task Adaptation.} The goal of the SemEval task was to generate a list of free-form paraphrases for a given noun compound, which explicate the implicit semantic relation between its constituent. To match with our other tasks, we cast the task as a binary classification problem where the input is a noun compound NC and a paraphrase p, and the goal is to predict whether p is a correct description of NC. 

The positive examples for an NC w$_1$ w$_2$ are trivially derived from the original data by sampling up to 5 of its paraphrases and creating a (NC, p, TRUE) example for each paraphrase p. The same number of negative examples is then created using negative sampling from the paraphrase templates of other noun compounds w'$_1$ w$_2$ and w$_1$ w'$_2$ in the dataset that share a constituent word with NC. For example, ``\textit{replacement part bought for body}'' is a negative example constructed from the paraphrase template ``\textit{replacement} [w$_2$] \textit{bought for} [w$_1$]'' which appeared for \textit{car part}. We require one shared constituent in order to form more fluent paraphrases (which would otherwise be easily classifiable as negative). To reduce the chances of creating negative examples which are in fact valid paraphrases, we only consider negative paraphrases whose verbs never occurred in the positive paraphrase set for the given NC. 

We add sentential contexts from Wikipedia, randomly selecting one sentence per example, and split the dataset lexically by head. 

\begin{table*}[!t]
    \scriptsize
    \centering
    \begin{tabular}{c l c c c c c c}
\toprule 
 & \textbf{training objective} &  \textbf{corpus (\#words)} &  \specialcell{\textbf{output}\\\textbf{dimension}} & \specialcell{\textbf{basic unit}} \\
\midrule
 \rowcolor{lightgray} \multicolumn{6}{c}{\textit{word embeddings}} \\ \midrule
 \smallsc{word2vec} & Predicting surrounding words & Google News (100B) & 300 & word \\
  \midrule
 \smallsc{GloVe} & Predicting co-occurrence probability & Wikipedia + Gigaword 5 (6B) & 300 & word \\
  \midrule
\smallsc{fastText} & Predicting surrounding words & \specialcell{Wikipedia + UMBC + statmt.org (16B)} & 300 & subword \\
  \midrule
 \rowcolor{lightgray} \multicolumn{6}{c}{\textit{contextualized word embeddings}} \\ \midrule
 \smallsc{ELMo} & Language model & 1B Word Benchmark (1B) & 1024 & character \\
   \midrule
 \specialcell{\smallsc{OpenAI GPT}}  & Language model & BooksCorpus (800M) & 768 & subword \\
   \midrule
 \smallsc{BERT} & Masked language model (Cloze)  & BooksCorpus + Wikipedia (3.3B) & 768 & subword \\
\bottomrule
\end{tabular}

    \caption{Architectural differences of the specific pre-trained representations used in this paper.}
    \label{tab:representations}
\end{table*}

\subsection{Adjective-Noun Attributes}  
\label{sec:an_attribute_selection}

\paragraph{Task Definition.} Given an adjective-noun composition AN in a sentence s, and an attribute AT, the goal is to determine whether AT is implicitly conveyed in AN. For example, the attribute \texttt{temperature} is conveyed in \textit{hot water}, but not in \textit{hot argument} (\texttt{emotionality}).

\paragraph{Data.} We use the HeiPLAS data set \cite{hartung2015distributional}, which contains 1,598 adjective-noun compositions annotated with their implicit attribute meaning. The data was extracted from WordNet and manually filtered. The label set consists of attribute synsets in WordNet which are linked to adjective synsets.

\paragraph{Task Adaptation.} Since the dataset is small and the number of labels is large (254), we recast the task as a binary classification task. The input to the task is an AN and a paraphrase created from the template ``[A] refers to the [AT] of [N]'' (e.g. ``\textit{loud refers to the volume of thunder}''). The goal is to predict whether the paraphrase is correct or not with respect to the given AN. 

We create up to 3 negative instances for each positive instance by replacing AT in another attribute that appeared with either A or N. For example, (\textit{hot argument}, \texttt{temperature}, False). To reduce the chance that the negative attribute is in fact a valid attribute for AN, we compute the Wu-Palmer similarity \cite{wu1994verbs} between the original and negative attribute, taking only attributes whose similarity to the original attribute is below a threshold (0.4). 


Similarly to the previous task, we attach a context sentence from Wikipedia to each example. Finally, we split the dataset lexically by adjective, i.e. if A N is in one set, there are no A N' examples in the other sets. 

\subsection{Identifying Phrase Type}
\label{sec:mwe_type}

The last task consists of multiple phrase types and addresses detecting both meaning shift and implicit meaning.

\paragraph{Task Definition.} The task is defined as sequence labeling to \texttt{BIO} tags. Given a sentence, each word is classified to whether it is part of a phrase, and the specific type of the phrase.

\paragraph{Data.} We use the STREUSLE corpus \cite[Supersense-Tagged Repository of English with a Unified Semantics for Lexical Expressions, ][]{schneider-smith:2015:NAACL-HLT}. The corpus contains texts from the web reviews portion of the English Web Treebank, along with various semantic annotations, from which we use the \texttt{BIO} annotations. Each token is labeled with a tag, a \texttt{B-X} tag marks the beginning of a span of type X, \texttt{I} occurs inside a span, and \texttt{O} outside of it. \texttt{B} labels mark specific types of phrase.\footnote{Sorted by frequency: noun phrase, weak (compositional) MWE, verb-particle construction, verbal idioms, prepositional phrase, auxiliary, adposition, discourse / pragmatic expression, inherently adpositional verb, adjective, determiner, adverb, light verb construction, non-possessive pronoun, full verb or copula, conjunction.}

\paragraph{Task Adaptation.} We are interested in a simpler version of the annotations. Specifically, we exclude the discontinuous spans (e.g. a span like ``\textit{turn the} [TV] \textit{off}'' would not be considered as part of a phrase). The corpus distinguishes between ``strong'' MWEs (fixed or idiomatic phrases) and ``weak'' MWEs (ad hoc compositional phrases). The weak MWEs are untyped, hence we label them as \texttt{COMP} (compositional). 

\section{Representations}
\label{sec:representations}
We experimented with 6 common word representations from two different families detailed below. Table~\ref{tab:representations} summarizes the differences between the pretrained models used in this paper.

\paragraph{Word Embeddings.} Word embedding models provide a fixed $d$-dimensional vector for each word in the vocabulary. Their training is based on the distributional hypothesis, according to which semantically-similar words tend to appear in the same contexts \cite{harris1954distributional}. \textbf{word2vec} \cite{mikolov2013efficient} can be trained with one of two objectives. We use the embeddings trained with the Skip-Gram objective which predicts the context words given the target word. \textbf{GloVe} \cite{pennington2014glove} learns word vectors with the objective of estimating the log-probability of a word pair co-occurrence. \textbf{fastText} \cite{bojanowski2017enriching} extends word2vec by adding information about subwords (bag of character n-grams). This is especially helpful in morphologically-rich languages, but can also help handling rare or misspelled words in English. 

\paragraph{Contextualized Word Embeddings} are functions computing dynamic word embeddings for words given their context sentence, largely addressing polysemy. They are pre-trained as general purpose language models using a large-scale unannotated corpus, and can later be used as a representation layer in downstream tasks (either fine-tuned to the task with the other model parameters or fixed). The representations used in this paper have multiple output layers. We either use only the last layer, which was shown to capture semantic information \cite{peters-EtAl:2018:N18-1}, or learn a task-specific scalar mix of the layers (see Section~\ref{sec:results}). 

\textbf{ELMo} \cite[Embeddings from Language Models,][]{peters-EtAl:2018:N18-1} are obtained by learning a character-based language model using a deep biLSTM \cite{graves2005framewise}. Working at the character-level allows using morphological clues to form robust representations for out-of-vocabulary words, unseen in training. The \textbf{OpenAI GPT} \cite[Generative Pre-Training,][]{radford2018improving} has a similar training objective, but the underlying encoder is a transformer \cite{vaswani2017attention}. It uses subwords as the basic unit, employing bytepair encoding. Finally, \textbf{BERT} \cite[Bidirectional Encoder Representations from Transformers,][]{devlin2018bert} is also based on the transformer, but it is bidirectional as opposed to left-to-right as in the OpenAI GPT, and the directions are dependent as opposed to ELMo's independently trained left-to-right and right-to-left LSTMs. It also introduces a somewhat different objective called ``masked language model'': during training, some tokens are randomly masked, and the objective is to restore them from the context. 

\section{Classification Models}
\label{sec:models}
We implemented minimal ``Embed-Encode-Predict'' models that use the representations from Section~\ref{sec:representations} as inputs, keeping them fixed during training. The rationale behind the model design was to keep them uniform for easy comparison between the representations, and make them simple so that the model's success can be attributed directly to the input representations. 

\paragraph{Embed.} We use the embedding model to embed each word in the sentence $s = w_1 ... w_n$, obtaining:

\begin{equation}
 \vec{v}_1, ..., \vec{v}_n = \operatorname{Embed}(s)   
\label{eq:embed_sent}
\end{equation}

\noindent where $\vec{v}_i$ stands for the word embedding of word $w_i$, which may be computed as a function of the entire sentence (in the case of contextualized word embeddings). 

Depending on the specific task, we may have another input $w'_1, ..., w'_l$ to embed separately from the sentence: the paraphrases in the NC Relations and AN Attributes tasks, and the target word in the NC Literality task (to obtain an out-of-context representation of the target word). We embed this extra input as follows:

\begin{equation}
 \vec{v'}_1, ..., \vec{v'}_l = \operatorname{Embed}(w'_1, ..., w'_l)   
\label{eq:embed_extra}
\end{equation}

\begin{table*}[!t]
    \scriptsize
    \centering
    \begin{tabular}{c c l}
\toprule
\textbf{Task} & \textbf{Agreement} & \textbf{Example Question} \\ 
\midrule
\textbf{VPC Classification} & 84.17\% & 
\specialcellleft{\textit{I feel there are others far more suited to \textbf{take on} the responsibility.}\\
What is the verb in the highlighted span? (take/take on)} \\
\midrule
\textbf{LVC Classification} & 83.78\% & 
\specialcellleft{
\textit{Jamie \textbf{made a decision} to drop out of college.}\\
Mark all that apply to the highlighted span in the given context:\\ 
1. It describes an action of ``\textit{making something}'', in the common meaning of ``\textit{make}''.\\
2. The essence of the action is described by ``\textit{decision}''. \\
3. The span could be rephrased without ``\textit{make}'' but with a verb like ``\textit{decide}'',\\without changing the meaning of the sentence.\\
}
\\
\midrule
\textbf{NC Literality} & 80.81\% & 
\specialcellleft{\textit{He is driving down memory \textbf{lane} and reminiscing about his first love.}\\
Is ``lane'' used literally or non-literally? (literal/non literal)} \\
\midrule
\textbf{NC Relations} & 86.21\% & 
\specialcellleft{\textit{Strawberry shortcakes were held as celebrations of the \textbf{summer fruit} harvest.}\\
Can ``summer fruit'' be described by ``fruit that is ripe in the summer''? (yes/no)} \\
\midrule
\textbf{AN Attributes} & 86.42\% & 
\specialcellleft{\textit{Send my \textbf{warm} regards to your parents.}\\Does ``warm'' refer to temperature? (yes/no)} \\
\bottomrule
\end{tabular}

    \caption{The worker agreement (\%) and the question(s) displayed to the workers in each annotation task.}
    \label{tab:annotation}
\end{table*}

\paragraph{Encode.} We encode the embedded sequences $\vec{v}_1, ..., \vec{v}_n $ and $\vec{v'}_1, ..., \vec{v'}_l$ using one of the following 3 encode variants. As opposed to the pre-trained embeddings, the encoder parameters are updated during training to fit the specific task.

\begin{itemize}[noitemsep,leftmargin=*]
    \item \texttt{biLM}: Encoding the embedded sequence using a biLSTM with a hidden dimension $d$, where $d$ is the input embedding dimension: 

    \begin{equation}
     \vec{u}_1, ..., \vec{u}_n = \operatorname{biLSTM}(\vec{v}_1, ..., \vec{v}_n)
    \label{eq:encode_bilm}
    \end{equation}
    
    \item \texttt{Att}: Encoding the embedded sequence using self-attention. Each word is represented as the concatenation of its embedding and a weighted average over other words in the sentence:
    
    \begin{equation}
     \vec{u}_i = [\vec{v}_i ; \sum_{j=1}^{n} a_{i, j} \cdot \vec{v}_j ]
    \label{eq:encode_att}
    \end{equation}
    
    \noindent The weights $a_i$ are computed by applying dot-product between $\vec{v}_i$ and every other word, and normalizing the scores using softmax:
    
    \begin{equation}
     \vec{a}_i = \operatorname{softmax}(\vec{v}_i^T \cdot \vec{v})
    \label{eq:encode_none}
    \end{equation}
    
    \item \texttt{None}: In which we don't encode the embedded text, but simply define:

    \begin{equation}
     \vec{u}_1, ..., \vec{u}_n = \vec{v}_1, ..., \vec{v}_n
    \label{eq:noenc_sent}
    \end{equation}
    
\end{itemize}

\noindent For all encoder variants, $\vec{u}_i$ stands for the vector representing $w_i$, which is used as input to the classifier. 

\paragraph{Predict.} We represent a span by concatenating its end-point vectors, e.g. $\vec{u}_{i...i+k} = [\vec{u}_i ; \vec{u}_{i+k}]$ is the target span vector of $w_i, ..., w_{i+k}$. In tasks which require a second span, we similarly compute $\vec{u'}_{1...l}$, representing the encoded span $w'_1, ..., w'_l$ (e.g. the paraphrase in NC relations). The input to the classifier is the concatenation of $\vec{u}_{i...i+k}$, and, when applicable, the additional span vector $\vec{u'}_{1...l}$. In the general case, the input to the classifier is:

\begin{equation}
\vec{x} = [\vec{u}_i ; \vec{u}_{i+k} ; \vec{u'}_1 ; \vec{u'}_l]
\label{eq:span_embedding}
\end{equation}

\noindent where each of $\vec{u}_{i+k}$, $\vec{u'}_1$, and $\vec{u'}_{l}$ can be empty vectors in the cases of single word spans or no additional inputs.

The classifier output is defined as:

\begin{equation}
\vec{o} = \operatorname{softmax}(W \cdot \operatorname{ReLU}(\operatorname{Dropout}(h(\vec{x})))) 
\label{eq:classifier}
\end{equation}

\noindent where $h$ is a 300-dimensional hidden layer, the dropout probability is 0.2, $W \in \mathcal{R}^{k \times 300}$, and $k$ is the number of class labels for the specific task. 

\paragraph{Implementation Details.} We implemented the models using the AllenNLP library \cite{Gardner2017AllenNLP} which is based on the PyTorch framework \cite{paszke2017automatic}. We train them for up to 500 epochs, stopping early if the validation performance doesn't improve in 20 epochs.

The phrase type model is a sequence tagging model that predicts a label for each embedded (potentially encoded) word $w_i$. During decoding, we enforce a single constraint that requires that a \texttt{B-X} tag must precede \texttt{I} tag(s).

\section{Baselines}
\label{sec:baselines}
\begin{table*}[!t]
    \scriptsize
    \centering
    \begin{tabular}{l c c c c c c}
\toprule
\textbf{Model Family} & \texttt{VPC} & \texttt{LVC} & \texttt{NC} & \texttt{NC} & \texttt{AN} & \texttt{Phrase} \\ 
& \texttt{Classification} & \texttt{Classification} & \texttt{Literality} & \texttt{Relations} & \texttt{Attributes} & \texttt{Type} \\
\toprule
& $Acc$ & $Acc$ & $Acc$ & $Acc$ & $Acc$ & $F_1$ \\ 
\textbf{Majority Baselines} & $23.6$ & $43.7$ & $72.5$ & $50.0$ & $50.0$ & $26.6$ \\ 
\textbf{Word Embeddings} & $60.5$ & $74.6$ & $80.4$ & $51.2$ & $53.8$ & $44.0$ \\
\textbf{Contextualized} & $\mathbf{90.0}$ & $\mathbf{82.5}$ & $\mathbf{91.3}$ & $\mathbf{54.3}$ & $\mathbf{65.1}$ & $\mathbf{64.8}$ \\
\textbf{Human} & $93.8$ & $83.8$ & $91.0$ & $77.8$ & $86.4$ & - \\ 
\bottomrule
\end{tabular}

    \caption{Summary of the best performance of each family of representations on the various tasks. The evaluation metric is accuracy except for the phrase type task in which we report span-based $F_1$ score, excluding \texttt{O} tags.}
    \label{tab:result_summary}
\end{table*}

\subsection{Human Performance}
\label{sec:human_performance}

The human performance on each task can be used as a performance upper bound which shows both the inherent ambiguity in the task, as well as the limitations of the particular dataset. We estimated the human performance on each task by sampling and re-annotating 100 examples from each test set. 

The annotation was carried out in Amazon Mechanical Turk. We asked 3 workers to annotate each example, taking the majority label as the final prediction. To control the quality of the annotations, we required that workers must have an acceptance rate of at least 98\% on at least 500 prior HITs, and had them pass a qualification test. 

In each annotation task, we showed the workers the context sentence with the target span highlighted, and asked them questions regarding the target span as exemplified in Table~\ref{tab:annotation}. In addition to the possible answers given in the table, annotators were always given the choice of ``I can't tell'' or ``the sentence does not make sense''. 

Of all the annotation tasks, the LVC classification task was more challenging and required careful examination of the different criteria for LVCs. In the example given in Table~\ref{tab:annotation} with the candidate LVC ``\textit{make a decision}'', we considered a worker's answer as positive (LVC) either if: 1) the worker indicated that ``\textit{make a decision}'' does not describe an action of ``\textit{making something}'' AND that the essence of ``\textit{make a decision}'' is in the word ``\textit{decision}''; or 2) if the worker indicated that ``\textit{make a decision}'' can be replaced in the given sentence with ``\textit{decide}'' without changing the meaning of the sentence. The second criterion was given in the original guidelines of \newcite{tu-roth:2011:MWE}. The replacement verb ``\textit{decide}'' was selected as it is linked to ``\textit{decision}'' in WordNet in the derivationally-related relation. 

We didn't compute the estimated human performance on the phrase type task, which is more complicated and requires expert annotation.

\subsection{Majority Baselines}
\label{sec:majority_baselines}

We implemented three majority baselines: 

\begin{itemize}[noitemsep,leftmargin=*]
    \item Majority$_{ALL}$ is computed by assigning the most common label in the training set to all the test items. Note that the label distribution may be different between the train and test sets, resulting in accuracy $< 50\%$ even for binary classification tasks.
    \item Majority$_1$ assigns for each test item the most common label in the training set for items with the same first constituent. For example, in the VPC classification task, it classifies \textit{get through} as positive in all its contexts because the verb \textit{get} appears in more positive than negative examples.
    \item Majority$_2$ symmetrically assigns the label according to the last (typically second) constituent. 
\end{itemize}

\section{Results}
\label{sec:results}
\begin{table*}[!t]
    \scriptsize
    \centering
    \begin{tabular}{p{1.65cm} 
>{\centering\arraybackslash}p{0.7cm} >{\centering\arraybackslash}p{0.7cm} 
>{\centering\arraybackslash}p{0.7cm} >{\centering\arraybackslash}p{0.7cm} 
>{\centering\arraybackslash}p{0.4cm} >{\centering\arraybackslash}p{1cm} 
>{\centering\arraybackslash}p{0.35cm} >{\centering\arraybackslash}p{0.35cm} 
>{\centering\arraybackslash}p{0.35cm} >{\centering\arraybackslash}p{0.35cm} 
>{\centering\arraybackslash}p{0.35cm} >{\centering\arraybackslash}p{0.6cm}}
\toprule
\textbf{Model} & 
\multicolumn{2}{c}{\specialcell{\smalltt{VPC} \\ \smalltt{Classification}}} & 
\multicolumn{2}{c}{\specialcell{\smalltt{LVC} \\ \smalltt{Classification}}} & 
\multicolumn{2}{c}{\specialcell{\smalltt{NC} \\ \smalltt{Literality}}} & 
\multicolumn{2}{c}{\specialcell{\smalltt{NC} \\ \smalltt{Relations}}} & 
\multicolumn{2}{c}{\specialcell{\smalltt{AN} \\ \smalltt{Attributes}}} & 
\multicolumn{2}{c}{\specialcell{\smalltt{Phrase} \\ \smalltt{Type}}}\\ 
\toprule
& \tiny{Layer} & \tiny{Encoding} & \tiny{Layer} & \tiny{Encoding} & \tiny{Layer} & \tiny{Encoding} & \tiny{Layer} & \tiny{Encoding} & \tiny{Layer} & \tiny{Encoding} & \tiny{Layer} & \tiny{Encoding} \\
\toprule
\textbf{ELMo} & \smalltt{All} & \smalltt{Att} & \smalltt{All} & \smalltt{biLM} & \smalltt{All} & \smalltt{Att}/\smalltt{None} & \textbf{\smalltt{Top}} & \textbf{\smalltt{biLM}} & \smalltt{All} & \smalltt{None} & \smalltt{All} & \smalltt{biLM} \\
\textbf{OpenAI GPT} & \smalltt{All} & \smalltt{None} & \textbf{\smalltt{Top}} & \textbf{\smalltt{Att}/\smalltt{None}} & \smalltt{Top} & \smalltt{None} & \smalltt{All} & \smalltt{biLM} & \smalltt{Top} & \smalltt{None} & \smalltt{All} & \smalltt{biLM} \\
\textbf{BERT} & \textbf{\smalltt{All}} & \textbf{\smalltt{Att}} & \smalltt{All} & \smalltt{biLM} & \textbf{\smalltt{All}} & \textbf{\smalltt{Att}} & \smalltt{All} & \smalltt{None} & \textbf{\smalltt{All}} & \textbf{\smalltt{None}} &  \textbf{\smalltt{All}} & \textbf{\smalltt{biLM}} \\ 
\bottomrule
\end{tabular}

    \caption{The best setting (layer and encoding) for each contextualized word embedding model on the various tasks. Bold entries are the best performers on each task.}
    \label{tab:results_contextualized}
\end{table*}

\begin{table*}[!t]
    \scriptsize
    \centering
    \begin{tabular}{l c c c c c c}\toprule
\textbf{Model} & 
\specialcell{\texttt{VPC} \\ \texttt{Classification}} & 
\specialcell{\texttt{LVC} \\ \texttt{Classification}} & 
\specialcell{\texttt{NC} \\ \texttt{Literality}} & 
\specialcell{\texttt{NC} \\ \texttt{Relations}} & 
\specialcell{\texttt{AN} \\ \texttt{Attributes}} & 
\specialcell{\texttt{Phrase} \\ \texttt{Type}}\\ 
\toprule
\textbf{word2vec} & \smalltt{biLM} & \smalltt{Att} & \smalltt{biLM}/\smalltt{Att} & \smalltt{None} & - & \smalltt{None}/\smalltt{biLM} \\
\textbf{GloVe} & \textbf{\smalltt{biLM}} & \textbf{\smalltt{Att}} & \textbf{\smalltt{Att}} & \smalltt{biLM} & - & \textbf{\smalltt{biLM}} \\
\textbf{fastText} & \smalltt{Att} & \smalltt{biLM} & \smalltt{biLM} & \textbf{\smalltt{biLM}} & \textbf{\smalltt{Att}} & \smalltt{biLM} \\
\bottomrule
\end{tabular}

    \caption{The best encoding for each word embedding model on the various tasks. Bold entries are the best performers on each task. Dash marks no preference.}
    \label{tab:results_embeddings}
\end{table*}

Table~\ref{tab:result_summary} displays the best performance of each model family on the various tasks. 

\paragraph{Representations.} The general trend across tasks is that the performance improves from the majority baselines through word embeddings and to the contextualized word representations, with a large gap in some of the tasks. Among the contextualized word embeddings, BERT performed best on 4 out of 6 tasks, with no consistent preference between ELMo and the OpenAI GPT. The best 
word embedding representations were GloVe (4/6) followed by fastText (2/6). 

\paragraph{Phenomena.} The gap between the best model performance (achieved by the contextualized representations) and the estimated human performance varies considerably across tasks. The best performance in NC Literality is on par with human performance, and only a few points short of that in VPC Classification and LVC Classification. This is an evidence for the utility of contextualized word embeddings in detecting meaning shift, which has positive implications for the yet unsolved problem of detecting MWEs. 

Conversely, the gap between the best model and the human performance is as high as 23.5 and 21.3 points in NC Relations and AN Attributes, respectively, suggesting that tasks requiring revealing implicit meaning are more challenging to the existing representations.

\paragraph{Layer.} Table~\ref{tab:results_contextualized} elaborates on the best setting for each representation on the various tasks. In most cases, there was a preference to learning a scalar mix of the layers rather than using only the top layer. We extracted the learned layer weights for each of the \texttt{All} models, and found that the model usually learned a balanced mix of the top and bottom layers. 

\paragraph{Encoding.} We did not find one encoding setting that performed best across tasks and contextualized word embedding models. Instead, it seems that tasks related to meaning shift typically prefer \texttt{Att} or no encoding, while tasks related to implicit meaning performed better with either \texttt{biLM} or no encoding. 

When it comes to word embedding models, Table~\ref{tab:results_embeddings} shows that \texttt{biLM} was preferred more often. This is not surprising. A contextualized word embedding of a certain word is, by definition, already aware of the surrounding words, obviating the need for a second layer of order-aware encoding. A word embedding based model, on the other hand, must rely on a biLSTM to learn the same. 

\paragraph{} Finally, the best settings on the Phrase Type task use \texttt{biLM} across representations. It may suggest that predicting a label for each word can benefit from a more structured modelling of word order. Looking into the errors made by the best model (BERT+All+biLM) reveals that most of the errors were predicting \texttt{O}, i.e. missing the occurrence of a phrase. With respect to specific phrase types, near perfect performance was achieved among the more syntactic categories. Specifically, auxiliary (``\textit{Did they think we were [going to] feel lucky to get any reservation at all?}''), adverbs (``\textit{any longer}''), and determiners (``\textit{a bit}''). 
In accordance to the VPC Classification task, the VPC label achieved 85\% accuracy. 10\% were missed (classified as \texttt{O}) and 5\% were confused with a ``weak'' MWE. Two of the more difficult types were ``weak'' MWEs (which are judged as more compositional and less idiomatic) and idiomatic verbs. The former achieved accuracy of 22\% (68\% were classified as \texttt{O}) and the latter only 8\% (62\% were classified as \texttt{O}). Overall it seems that the model relied mostly on syntactic cues, failing to recognize semantic subtleties such as idiomatic meaning and level of compositionality.

\section{Analysis}
\label{sec:analysis}
We focus on the contextualized word embeddings, and look into the representations they provide.

\begin{figure}[!t]
    \centering
    \vspace{6pt}
    \begin{tabular}{|c|c|}
    \hline
    \includegraphics[trim={50pt 33pt 20pt 5pt},clip,width=0.21\textwidth]{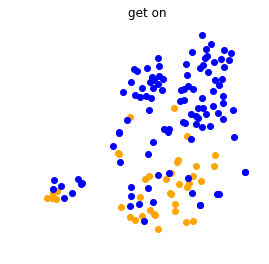} &
    \includegraphics[trim={55pt 33pt 20pt 5pt},clip,width=0.21\textwidth]{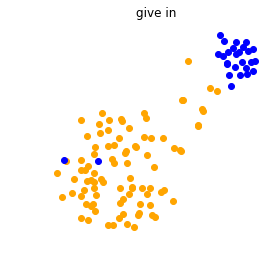}
    \\
    \hline
    \includegraphics[trim={50pt 33pt 20pt 5pt},clip,width=0.21\textwidth]{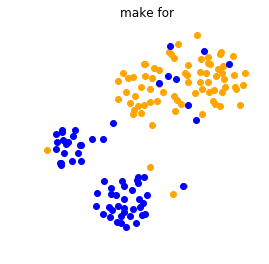} &
    \includegraphics[trim={55pt 33pt 20pt 5pt},clip,width=0.21\textwidth]{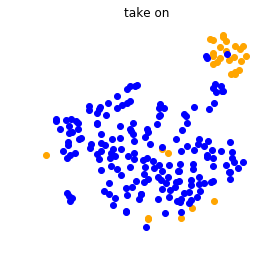} \\
    \hline
    \end{tabular}
    \caption{t-SNE projection of BERT representations of verb-preposition candidates for VPC. Blue (dark) points are positive examples and orange (light) points are negative.}
    \label{fig:bert_vpc}
\end{figure}

\begin{table*}[!t]
    \scriptsize
    \centering
    \begin{tabular}{p{0.075\textwidth}p{0.03\textwidth}p{0.075\textwidth}p{0.03\textwidth}p{0.075\textwidth}p{0.03\textwidth}p{0.01\textwidth}p{0.075\textwidth}p{0.03\textwidth}p{0.075\textwidth}p{0.03\textwidth}p{0.075\textwidth}p{0.03\textwidth}}
\toprule
\multicolumn{2}{c}{\smalltt{ELMo}} & \multicolumn{2}{c}{\specialcellleft{\smalltt{OpenAI GPT}}} & \multicolumn{2}{c}{\smalltt{BERT}} & & \multicolumn{2}{c}{\smalltt{ELMo}} & \multicolumn{2}{c}{\specialcellleft{\smalltt{OpenAI GPT}}} & \multicolumn{2}{c}{\smalltt{BERT}} \\ 
\toprule
\multicolumn{6}{l}{\specialcellleft{The Queen and her husband were on a train [trip]$_L$ from \\ Sydney to Orange.}}  &  & 
\multicolumn{6}{l}{\specialcellleft{Creating a guilt [trip]$_N$ in another person may be considered to \\ be psychological manipulation in the form of punishment for a \\ perceived transgression.}}  \\ 
\midrule
\textbf{ride} & 1.24\% & to & 0.02\% & \textbf{travelling} & 19.51\% & & tolerance & 0.44\% & that & 0.03\% & \textbf{reaction} & 8.32\% \\ \textbf{carriage} & 1.02\% & \textbf{headed} & 0.01\% & \textbf{running} & 8.20\% & & \textbf{fest} & 0.23\% & so & 0.02\% & \textbf{feeling} & 8.17\% \\ \textbf{journey} & 0.73\% & \textbf{heading} & 0.01\% & \textbf{journey} & 7.57\% & & avoidance & 0.16\% & \textbf{trip} & 0.01\% & attachment & 8.12\% \\ \textbf{heading} & 0.72\% & that & 0.009\% & \textbf{going} & 6.56\% & & onus & 0.15\% & he & 0.01\% & \textbf{sensation} & 4.73\% \\ carrying & 0.39\% & and & 0.005\% & \textbf{headed} & 5.75\% & & association & 0.14\% & she & 0.01\% & note & 3.31\% \\ \midrule
\multicolumn{6}{l}{\specialcellleft{Richard Cromwell so impressed the king with his valour, that \\ he was given a [diamond]$_L$ ring from the king's own finger.}}  &  & 
\multicolumn{6}{l}{\specialcellleft{She became the first British monarch to celebrate a [diamond]$_N$\\ wedding anniversary in November 2007.}}  \\ 
\midrule
\textbf{diamond} & 0.23\% & and & 0.01\% & \textbf{silver} & 15.99\% & & customary & 0.20\% & new & 0.11\% & royal & 1.58\% \\
\textbf{wedding} & 0.19\% & of & 0.01\% & \textbf{gold} & 14.93\% & & royal & 0.17\% & british & 0.02\% & 1912 & 1.23\% \\ 
\textbf{pearl} & 0.18\% & to & 0.01\% & \textbf{diamond} & 13.18\% & & sacrifice & 0.15\% & victory & 0.01\% & recent & 1.10\% \\ 
knighthood & 0.16\% & ring & 0.01\% & \textbf{golden} & 12.79\% & & 400th & 0.13\% & french & 0.01\% & 1937 & 1.08\% \\ 
hollow & 0.15\% & in & 0.01\% & new & 4.61\% & & \textbf{10th} & 0.13\% & royal & 0.01\% & 1902 & 1.08\% \\ \midrule
\multicolumn{6}{l}{\specialcellleft{China is attempting to secure its future [oil]$_L$ share and establish \\ deals with other countries.}}  &  & 
\multicolumn{6}{l}{\specialcellleft{Put bluntly, I believe you are a snake [oil]$_N$ salesman, a\\narcissist that would say anything to draw attention to himself.}}  \\ \midrule
beyond & 0.44\% & in & 0.01\% & \textbf{market} & 98.60\% & & auto & 0.52\% & in & 0.05\% & \textbf{oil} & 32.5\% \\ engagement & 0.44\% & and & 0.01\% & \textbf{export} & 0.45\% & & egg & 0.42\% & and & 0.01\% & pit & 2.94\% \\ \textbf{market} & 0.34\% & for & 0.01\% & \textbf{trade} & 0.14\% & & hunter & 0.42\% & that & 0.01\% & bite & 2.65\% \\ \textbf{nuclear} & 0.33\% & \textbf{government} & 0.01\% & \textbf{trading} & 0.09\% & & consummate & 0.39\% & of & 0.01\% & skin & 2.36\% \\ majority & 0.29\% & \textbf{supply} & 0.01\% & \textbf{production} & 0.04\% & & \underline{rogue} & 0.37\% & \underline{charmer} & 0.008\% & jar & 2.23\% \\ 
\bottomrule
\end{tabular}

    \caption{Top substitutes for a target word in literal (left) and non-literal (right) contexts, along with model scores. Bold words are words judged reasonable (not necessarily meaning preserving) in the given context, and underlined words are suitable substitutes for the entire noun compound, but not for a single constituent.}
    \label{tab:predictions}
\end{table*}

\subsection{Meaning Shift}
\label{sec:meaning_shift_analysis}

\paragraph{Does the representation capture VPCs?} The best performer on the VPC Classification task was the BERT+All+Att. To get a better idea of the signal that BERT contains for VPCs, we chose several ambiguous verb-preposition pairs in the dataset. We define a verb-preposition pair as ambiguous if it appeared in at least 8 examples as a VPC and at least 8 examples as a non-VPC. For a given pair we computed the BERT representations of the sentences in which it appears in the dataset, and, similarly to the model, we represented the pair as the concatenation of its word vectors. In each vector we averaged the layers using the weights learned by the model. Finally, we projected the computed vectors into 2D space using t-SNE \cite{maaten2008visualizing}. Figure~\ref{fig:bert_vpc} demonstrates 4 example pairs. The other pairs we plotted had similar t-SNE plots, confirming that the signal for separating different verb usages comes directly from BERT.  

\paragraph{Non-literality as a rare sense.} \newcite{nunberg1994idioms} considered some non-literal compounds as ``idiosyncratically decomposable'', i.e. which can be decomposed to possibly rare senses of their constituents, as in considering \textit{bee} to have a sense of ``competition'' in \textit{spelling bee} and \textit{crocodile} to stand for ``manipulative'' in \textit{crocodile tears}. Using this definition, we could possibly use the NC literality data for word sense induction, in which recent work has shown that contextualized word representations are successful \cite{stanovsky-hopkins:2018:EMNLP,amrami-goldberg:2018:EMNLP}.  We are interested in testing not only whether the contextualized models are capable of detecting rare senses induced by non-literal usage, which we have confirmed in Section~\ref{sec:results}, but whether they can also model these senses. To that end, we sample target words that appear in both literal and non-literal examples, and use each contextualized word embedding model as a language model to predict the best substitutes of the target word in each context. Table~\ref{tab:predictions} exemplifies some of these predictions. 

Bold words are words judged reasonable in the given context, even if they don't have the exact same meaning as the target word. It is apparent that there are more reasonable substitutes for the literal examples, across models (left part of the table), but BERT performs better than the others. The OpenAI GPT shows a clear disadvantage of being uni-directional, often choosing a substitute that doesn't go well with the next word (``\textit{a train \textbf{to} from}''). 

The success is only partial among non-literal examples. While some correct substitutes are predicted for \textit{(guilt) trip}, the predictions are much worse for the other examples. The meaning of diamond in \textit{diamond wedding} is ``60th'', and ELMo makes the closest prediction, \textit{10th} (which would make it ``\textit{tin wedding}''). \textit{400th} is a borderline prediction, because it is also an ordinal number, but an unreasonable one in the context of years of marriage. 

Finally, the last example \textit{snake oil} is unsurprisingly a difficult one, possibly ``non-decomposable'' \cite{nunberg1994idioms}, as both constituents are non-literal. Some predicted substitutes, \textit{rogue} and \textit{charmer}, are valid replacements for the entire noun compound (e.g. ``\textit{you are a rogue salesman}''). Others go well with the literal meaning of \textit{snake} creating phrases denoting concepts which can indeed be sold (\textit{snake egg}, \textit{snake skin}). 

Overall, while contextualized representations excel at detecting shifts from the common meanings of words, their ability to obtain meaningful representations for such rare senses is much more limited. 

\subsection{Implicit Meaning}
\label{sec:implicit_meaning_analysis}

The performance of the various models on the tasks that involve revealing implicit meaning are substantially worse than on the other tasks. In NC Relations, ELMo performs best with the biLM-encoded model using only the top layer of the representation, surpassing the majority baseline by only 4.3 points in accuracy. The best performer in AN Attributes is BERT, with no encoding and using all the layers, achieving accuracy of 65.1\%, well above the majority baseline (50\%). 

\begin{table}[!t]
    \scriptsize
    \centering
    \begin{tabular}{lcc}
\hline
&   \textbf{NC Relations} &   \textbf{AN Attributes} \\
\hline
      \texttt{Majority}      &          50.0 &           50.0    \\
 \texttt{-Phrase} &          50.0 &           55.66 \\
 \texttt{-Context}  &          45.06 &           63.21 \\
  \texttt{-(Context+Phrase)}   &          45.06 &           59.43 \\
   \texttt{Full Model}     &          54.3  &           65.1  \\
\hline
\end{tabular}
    \caption{Accuracy scores of ablations of the phrase, context sentence, and both features from the best models in the NC Relations and AN Attributes tasks (ELMo+Top+biLM and BERT+All+None, respectively).}
    \label{tab:implicit_meaning_ablations}
\end{table}

We are interested in finding out where the knowledge of the implicit meaning originates. Is it encoded in the phrase representation itself, or does it appear explicitly in the context sentences? Finally, could it be that the performance gap from the majority baseline is due to the models learning to recognize which paraphrases are more probable than others, regardless of the phrase itself?

To try answer this question, we performed ablation tests for each of the tasks, using the best performing setting for each (ELMo+Top+biLM for NC Relations and BERT+All+None for AN Attributes). We trained the following models (-X signifies the ablation of the X feature):

\begin{enumerate}
    \item \textbf{-Phrase}: where we mask the phrase in its context sentence, e.g. replacing ``\textit{Today, the house has become a wine bar or bistro called Barokk}'' with ``\textit{Today, the house has become a \textbf{something} or bistro called Barokk}''. Success in this setting may indicate that the implicit information is given explicitly in some of the context sentences.\footnote{A somewhat similar phenomenon was recently reported by \newcite{senaldi2019neural}. Their model managed to distinguish idioms from non-idioms, but their ablation study showed the model was in fact learning to distinguish between abstract contexts (in which idioms tend to appear) and concrete ones.}
    
    \item \textbf{-Context}: the out-of-context version of the original task, in which we replace the context sentence by the phrase itself, as in setting it to ``\textit{wine bar}''. Success in this setting may indicate that the phrase representation contains this implicit information.

    \item \textbf{-(Context+Phrase)}: in which we omit the context sentence altogether, and provide only the paraphrase, as in ``\textit{bar where people buy and drink wine}''. Success in this setting may indicate that negative sampled paraphrases form sentences which are less probable in English.
\end{enumerate}

Table~\ref{tab:implicit_meaning_ablations} shows the results of this experiment. A first observation is that the full model performs best on both tasks, suggesting that the model captures implicit meaning from various sources. In the NC Relations, all variants perform on par or worse than the majority baseline, achieving a few points less than the full model. In the AN Attributes task it is easier to see that the phrase (AN) is important for the classification, while the context is secondary. 

\section{Related Work}
\label{sec:related_work}
\paragraph{Probing Tasks.} One way to test whether dense representations capture a certain linguistic property is to design a probing task for this property, and build a model that takes the tested representation as an input. This kind of ``black box'' testing has become popular recently. \newcite{adi2016fine} studied whether sentence embeddings capture properties such as sentence length and word order. \newcite{conneau-EtAl:2018:Long} extended their work with a large number of sentence embeddings, and tested various properties at the surface, syntactic, and semantic levels. Others focused on intermediate representations in neural machine translation systems \cite[e.g][]{D16-1159,belinkov-EtAl:2017:Long,dalvi-EtAl:2017:I17-1,sennrich:2017:EACLshort}, or on specific linguistic properties such as agreement \cite{W18-5426}, and tense \cite{W18-5440}. 

More recently, \newcite{tenney2018what} and \newcite{liu2019linguistic} each designed a suite of tasks to test contextualized word embeddings on a broad range of sub-sentence tasks, including part-of-speech tagging, syntactic constituent labeling, dependency parsing, named entity recognition, semantic role labeling, coreference resolution, semantic proto-role, and relation classification. \newcite{tenney2018what} found that all the models produced strong representations for syntactic phenomena, but gained smaller performance improvements upon the baselines in the more semantic tasks. \newcite{liu2019linguistic} found that some tasks (e.g., identifying the tokens that comprise the conjuncts in a coordination construction) required fine-grained linguistic knowledge which was not available in the representations unless they were fine-tuned for the task. To the best of our knowledge, we are the first to provide an evaluation suite consisting of tasks related to lexical composition. 

\paragraph{Lexical Composition.} There is a vast literature on multi-word expressions in general \cite[e.g.][]{sag2002multiword,R11-1040}, and research focusing on noun compounds \cite[e.g.][]{nakov_2013,nastase2013semantic}, adjective-noun compositions \cite[e.g.][]{baroni-zamparelli:2010:EMNLP,boleda-et-al:2013:IWCS2013}, verb-particle constructions \cite[e.g.][]{baldwin2005deep,pichotta-denero:2013:EMNLP}, and light verb constructions \cite[e.g.][]{tu-roth:2011:MWE,chen2015english}.

In recent years, word embeddings have been used to predict the compositionality of phrases \cite{salehi-cook-baldwin:2015:NAACL-HLT,cordeiro-EtAl:2016:P16-1}, and to identify the implicit relation in adjective-noun compositions \cite{hartung-EtAl:2017:EACLlong} and in noun compounds \cite{surtani2015vsm,dima2016compositionality,shwartz-waterson:2018:N18-2,P18-1111}. 

\newcite{pavlick-callisonburch:2016:P16-1} created a simpler variant of the recognizing textual entailment task \cite[RTE, ][]{dagan2013recognizing} that tests whether an adjective-noun composition entails the noun alone and vice versa in a given context. They tested various standard models for RTE and found that the models performed poorly with respect to this phenomenon. To the best of our knowledge, contextualized word embeddings haven't been employed for tasks related to lexical composition yet.

\paragraph{Phrase Representations.} With respect to obtaining meaningful phrase representations, there is a prominent line of work in learning a composition function over pairs of words. \newcite{mitchell2010composition} suggested simple composition via vector arithmetics. \newcite{baroni-zamparelli:2010:EMNLP} and later \newcite{maillard-clark:2015:CoNLL} treated adjectival modifiers as functions that operate on nouns and change their meanings, and represented them as matrices. \newcite{zanzotto2010estimating} and \newcite{dinu-pham-baroni:2013:CVSC} extended this approach and composed any two words by multiplying each word vector by a composition matrix. These models start by computing the phrases' distributional representation (i.e. treating it as a single token) and then learning a composition function that approximates it. 

The main drawbacks of this approach are that it assumes compositionality and that it operates on phrases with a pre-defined number of words. Moreover, we can expect the resulting compositional vectors to capture properties inherited from the constituent words, but it is unclear whether they also capture new properties introduced by the phrase. For example, the compositional representation of \textit{olive oil} may capture properties like \textit{green} (from \textit{olive}) and \textit{fat} (from \textit{oil}), but would it also capture properties like \textit{expensive} (a result of the extraction process)?

Alternatively, other approaches were suggested for learning general phrase embeddings, either using direct supervision for paraphrase similarity \cite{wieting2015towards}, indirectly from an extrinsic task \cite{D12-1110}, or in an unsupervised manner by extending the word2vec objective \cite{poliak-EtAl:2017:EACLshort}. While they don't have constrains on the phrase length, these methods still suffer from the two other drawbacks above: they  assume that the meaning of the phrase can always be composed from its constituent meanings, and it is unclear whether they can incorporate implicit information and new properties of the phrase. We expected that contextualized word embeddings, which assign a different vector for a word in each given context, would address at least the first issue by producing completely different vectors to literal vs. non-literal word occurrences.

\section{Discussion and Conclusion}
\label{sec:conclusion}
We showed that contextualized word representations perform generally better than static word embeddings on tasks related to lexical composition. However, while they are on par with human performance in recognizing meaning shift, they are still far from that in revealing implicit meaning. This gap may suggest a limit on the information that distributional models currently provide about the meanings of phrases. 

Going beyond the distributional models, an approach to build meaningful phrase representations can get some inspiration from the way that humans process phrases. A study on how L2 learners process idioms found that the most common and successful strategies were inferring from the context (57\% success) and relying on the literal meanings of the constituent words (22\% success) \cite{cooper1999processing}. As opposed to distributional models that aim to learn from a large number of (possibly noisy and uninformative) contexts, the sentential contexts in this experiment were manually selected, and a follow up study found that extended contexts (stories) help the interpretation further \cite{asl2013impact}. The participants didn't simply rely on adjacent words or phrases, but also employed reasoning. For example, in the sentence ``\textit{Robert knew that he was robbing the cradle by dating a sixteen-year-old girl}'', the participants inferred that 16 is too young to date, combined it with the knowledge that \textit{cradle} is where a baby sleeps, and concluded that \textit{rob the cradle} means dating a very young person. This level of context modeling seems to be beyond the scope of current text representations. 

We expect that improving the ability of representations to reveal implicit meaning will require training them to handle this specific phenomenon. Our evaluation suite, the data and code will be made available. It is easily extensible, and may be used in the future to evaluate new representations for their ability to address lexical composition.

\section*{Acknowledgments}

This work was supported in part by an Intel ICRI-CI grant, the Israel Science Foundation grant 1951/17, the German Research Foundation through and the German-Israeli Project Cooperation (DIP, grant DA 1600/1-1). Vered is also supported by the Clore Scholars Programme (2017).

\bibliography{references}
\bibliographystyle{acl_natbib}

\end{document}